\documentclass[12pt]{article}
\usepackage[top=1in, bottom=1in, left=1in, right=1in]{geometry}
\usepackage{tabularx}
\usepackage{hyperref} 
\usepackage{pdfsync}
\usepackage{xcolor}

\usepackage{listings}
\usepackage{makecell}
\usepackage{booktabs}
\usepackage{graphicx}
\date{}
\usepackage{caption}
\usepackage{subcaption}
\usepackage{natbib}

\begin{document}

\title{Large Models of What? Mistaking Engineering Achievements for Human Linguistic Agency}

\author{%
  Abeba Birhane\\
  Mozilla Foundation and\\
  School of Computer Science and Statistics \\
  Trinity College Dublin, Ireland \\
  \and
  Marek McGann\\ 
  Department of Psychology\\
  Mary Immaculate College,
  Limerick, Ireland \\
  }

\maketitle

\begin{abstract}

In this paper we argue that key, often sensational and misleading, claims regarding linguistic capabilities of Large Language Models (LLMs) are based on at least two unfounded assumptions: the \textit{assumption of language completeness} and the \textit{assumption of data completeness}.  Language completeness assumes that a distinct and complete thing such as ``a natural language'' exists, the essential characteristics of which can be effectively and comprehensively modelled by an LLM. The assumption of data completeness relies on the belief that a language 
can be quantified and wholly captured by 
data. 
Work within the enactive approach to cognitive science makes clear that, rather than a distinct and complete thing, language is a means or way of acting. Languaging is not the kind of thing that can admit of a complete or comprehensive modelling. From an enactive perspective we identify three key characteristics of enacted language; \textit{embodiment}, \textit{participation}, and \textit{precariousness}, that are absent in LLMs, and likely incompatible in principle with current architectures. We argue 
that these absences imply that LLMs are not now and cannot in their present form be linguistic agents the way humans are. We illustrate the point in particular through the phenomenon of ``algospeak'', a recently described pattern of high-stakes human language activity in heavily controlled online environments. On the basis of these points, we conclude that sensational and misleading claims about LLM agency and capabilities emerge from a deep misconception of both what human language is and what LLMs are.

\end{abstract}

\section{Contrasting Agencies}

The current machine learning narrative is surrounded by extravagant claims,  
over-ethusiasm, and hype. The discourse around Language Language Models (LLMs) exemplifies its peak. Fascinated evangelists claim that these models are capable of ``understanding language''~\cite{wang2019superglue}, can ``store, combine, and reason about scientific knowledge''~\cite{taylor2022galactica}, are approaching Artificial General Intelligence (AGI), hold early ``sparks of AGI''~\cite{bubeck2023sparks}, that they are surpassing human capabilities~\cite{xi2023rise}, and may even be (or soon become) conscious~\cite{chalmers2023could,de2022google}.  

In order to evaluate these claims 
we place side by side on the one hand, what it is that LLMs do, and on the other hand, what human beings do when engaged in linguistic interaction. We find 
some significant differences, which we believe tend to undermine direct or `literal' comparisons between people and LLMs.

Hyperbolic claims  
surrounding LLMs often (mis)uses terms 
that are naturally applied to the experiences, capabilities, and characteristics of human beings~\cite{bender2020climbing,shanahan2022talking,mitchell2019artificial}. 
The continued use of these terms, where such discourse is not re-calibrated in line with the comparisons, gradually shifts the meanings of 
words like ``language" and ``understanding". 
The literal use of these terms in this context 
re-orients their meanings  
in line with what is instantiated by the machines, and by the systems in which these machines are and will be inserted as powerful artefacts. Mistaking the impressive engineering achievements of LLMs for the mastering of human language, language understanding, and linguistic acts has dire implications for various forms of social participation, human agency, justice  
and 
 policies surrounding them.

In the context of this special issue on the implications of an enactive perspective on 
human beings and technologies,  
we evaluate the relationship between human linguistic agency and the operations  
of LLMs;  
what these two things have in common, and how they differ. Comparing human linguistic practice to LLMs is itself problematic, given that 1) there is no standard or average human whose lingustic activities can be compared against that of  
LLMs
, and 2) the metrics and benchmarks used to evaluate the performance of LLMs are riddled with various issues~\cite{burnell2023rethink,meister2021language,aiyappa2023can}. Having said that, if it were the case that human language and the way  
LLMs operate have much in common, it will be reasonable to consider them two examples of the same phenomenon. 
In what follows, we argue that  
it is possible to offer generous interpretations of some aspects of LLM engineering to find parallels with human language learning. However, in the majority of key aspects of language learning and use, most specifically in the various kinds of linguistic agency exhibited by human beings, these small apparent comparisons do little to balance what are  much more deep-rooted contrasts.

In keeping with the scope of this special issue we consider human linguistic agency from an enactive perspective.  And we lean on academic literature, industry practices, and public discourse to draw our understanding and description of LLMs. The rest of the paper is structured as follows: Section~\ref{sec:two_conceptions}, contrasts two conceptions of language: the one instantiated in LLM engineering, and the one put forward by enactive cognitive science. Section~\ref{sec:comparisons} highlights 
what LLMs and people appear to have in common. In Section~\ref{sec:contrasts}, we contrast precarious embodied linguistic human participation 
with activities of LLMs. Section~\ref{sec:algospeak} delves into a recent phenomenon –- algospeak –- to illustrate linguistic agency in action and we conclude in Section~\ref{sec:conclusion}.

\section{Two Conceptions of Language}
\label{sec:two_conceptions}
The statistical and computational sciences behind the development of LLMs on the one hand, and enactive cognitive science on the other,  involve sharply distinct conceptions of what language is. In fact, the former rarely engages in rigorous conceptual understanding and analysis  
of language, but in engineering tools that imitate linguistic activity. This is 
a key point that underscores  
differences in values and goals between these different research communities \cite{chemeroLLMsDifferHuman2023}. As an analogy, artificial flight does not involve the kinds of things that are used to achieve flight by animals in non-human ecosystems. The goals of aeronautical engineers are not those of zoologists. Their methods and aims diverge accordingly.

The goals and aims of LLM engineers are not understanding human linguistic activity. Their goals relate to the production of language-like performance in text (and audio) production. This   
means 
they have little bearing on natural linguistic interaction. This in no way undermines the incredible engineering achievements of LLMs, but it does help us understand that we should no more believe that LLM output is the same phenomenon as human language than we believe that an Airbus 8320 tells us something important about hummingbird flight\footnote{It is important to note that the different professional goals of LLM engineers as compared with cognitive scientists do not imply a reduced ethical burden. In fact, given the almost certain ubiquity of LLM-produced text and generated speech in the coming years, not only is there an extended analogy with that of artificial flight, and the pollution of ecosystems, but additional considerations also apply. Additionally, language is a domain which is quintessentially human, and coherent speech production has, to this point in our history, been a strong positive indicator of the presence of ethical duties toward the producer (though the converse does not hold). In circumstances in which people become inured to the experience of dismissing LLMs as the interfaces to machines, corporate or otherwise, the risk of increasing the already horrendous dehumanisation inherent in much online and offline activity becomes great. How these risks should be mitigated should be a vibrant domain of discussion for the burgeoning field of LLM development, much more so than spurious concerns about existential risks to human civilisation.}.

These differences in aims and values between the cognitive scientific and engineering communities results in quite distinct understandings and assumptions   
of what language is, what counts as engaging in linguistic interaction, as well as widely differing views of how language relates to other aspects of being. 

\subsection{Large Models of What?}
In 1948, Claude Shannon wrote on the relation between language and entropy that: “Frequently the messages have meaning; that is they refer to or are correlated according to some system with certain physical or conceptual entities. These semantic aspects of communication are irrelevant to the engineering problem.”~\cite[p.3]{shannon1948mathematical}. Current large language models such as Gemini, Bard, Llama, Megatron Turing, Bloom, and the GPT variants, are an engineering endeavour in these terms, albeit much more sophisticated and enormous in scale compared with Shannon’s original concept. Shannon’s original idea of entropy, that what is being measured, represented, and manipulated is form and not meaning, remains relevant now for LLMs.

Layers of artificial neural network architectures, most notably transformer-based systems and diffusion-based approaches are initialised and trained using massive datasets, often text and  
text-image pairs, typically sourced from the World Wide Web (WWW) and condensed and filtered via numerous automated mechanisms. 
In transformer based learning, the data are broken down into ``tokens", typically a few characters in length, and the LLM develops a statistical representation of the relationship between billions of different tokens. While there are various ways in which the operation of these systems can be described, they are ultimately well-defined calculations across tokens, which results in a system which, when given an input, treats that input as the initial movements through a space of possible valid moves (token concatenations), and continues the path of those movements depending on the complexity of the model and the availability of computational power (with longer paths requiring much greater resources). In a sense, the LLM is comprised of a statistical model of the relationship between tokens in a dataset, and a pathfinding mechanism optimised to generate valid sequences of token concatenations using that model, typically displayed as text. In keeping with Shannon's comment above, we note that the processing of  datasets and the generation of output are engineering problems, word prediction or sequence extension grounded in the underlying distribition of previously processed text. The generated text need not necessarily adher to ``facts'' in the real world. 

For this reason, LLMs have been dubbed as Stochastic Parrots~\cite{bender2021dangers}, Bullshit Generators~\cite{McQuillan23,hicksChatGPTBullshit2024}, and The Great Pretender~\cite{Coldewey23}, amongst other terms. Though the output from such models often seems impressively meaningful, that apparent meaningfulness depends on the extent to which the meaning of a given word, phrase, or paragraph, can be represented by the relationships between tokens in the original dataset. For example, multilingual models trained on multilingual datasets show poor quality performance for languages outside the status quo~\cite{kreutzer2022quality,khanuja2021muril,wu2020all}. The language that an LLM represents is  
conceived by its engineers and developers as  something static and complete, which can be captured in the relationships between tokens. 

From this description we can see that claims regarding linguistic capabilities of LLMs depend on two implicit assumptions of  
language. The first is what we call the \textit{assumption of language completeness} - that there exists a ``thing", called a ``language" that is complete, stable,  quantifiable, and available for extraction from traces in the environment. The engineering problem then becomes how that ``thing” can be reproduced artificially.
The second assumption is the \textit{assumption of data completeness} - that all of the essential characteristics can be represented in the datasets that are used to initialise and ``train" the model in question. In other words, 
all of the essential characteristics of language use are assumed to be present within the relationships between tokens, which presumably would  
allow 
LLMs to effectively and comprehensively reproduce the ``thing" that is 
being modelled. 

Both of these assumptions are rejected by an enactive view of language, which sees it not as  a ``thing" to be captured by text data, 
but a practice in which to participate, whether that participation is through speech, written, sign, or other modality. In contrast with computational approaches'  
emphasis on form, the enactive approach to language recognises that what truly matters for language is its meaning (in this there are strong resonances with \cite{bender2020climbing}'s  
critical examination of the relationship between form and meaning in language model output). As such, the enactive approach to language starts not with tokens of verbal activity, but with the fundamental issue of agency, embodiment, precarity, and how meaning arises within situations where things matter to those involved.

\subsection{Doing Language}
Enactive descriptions of agency begin with a process of continuous precariousness -- a network of processes constituting a body in a constant flux of degradation for which engagement with and acting on the world provides some prospect of continuity through the managing of  conflicts and tensions that result in the degradation~\cite{beerTheoreticalFoundationsEnaction2023} .The logic of precarious but continuous dynamic organisation is one that arises in the domain of organic, bodily being, but is general across other domains.  
\cite{thompson2001radical} first outlined three such domains -- the organic (that of biological bodies), the sensorimotor (that of skilfully engaged bodies), and the intersubjective (that of interacting skilfully engaged bodies). These three domains can be conceived of as distinct, but nevertheless interact in inextricable ways. Organic needs can drive and constrain sensorimotor activity, while those skilful doings in turn affect and transform those biological dynamics. Similarly, the intersubjective is a coordination of skilful bodies interacting with one another, but this domain is constrained by organic and skill-relevant dynamics, while also imposing constraints in turn. Think of the ways in which societies (a collection of intersubjective phenomena) are substantially organised around food production and meal times, constrained by the biology of hunger and nutrition. Hunger and nutrition are in turn affected by social norms and standards around what kinds of foods are generally available (within the society's ``cuisine") and standardised meal times. These impacts occur over multiple timescales too, as we can see in such phenomena as available foods and cultural resources leading to rises and falls in prevalences of Type II diabetes~\cite{magliano2019trends} intersubjective phenomena driving and constraining biological ones.

Within an enactive perspective, agency generally, then, is driven by tensions of precariousness and risk. There are continual needs to diffuse a tension by, for instance, expending energy to move and act. These actions, however, necessarily introduce new tensions, such as a hunger arising from the expenditure of energy. \cite{di2018linguistic} provide an extensive characterisation of this inter-dependency between the organic, the sensorimotor, and the intersubjective, and extend this analysis of agency into the linguistic domain. \textit{Linguistic acts} are those which manage an inter-subjective tension -- a precariousness in the coordination between two or more people engaged together in a shared activity. Their analysis places great emphasis on the variety of ways in which such shared activities occur at multiple temporal and spatial scales, and that the resolution of a tension at one scale tends to introduce tensions at another.

There are two crucial implications for our understanding of language that follow from \cite{di2018linguistic}'s account. The \textit{first} is that we are always doing more than one thing. Linguistic actions are made within a nested set of contexts. When we encounter other people we are always already in some broad form of coordination with them in which we are participating. For instance, if we meet someone for the first time at a job interview we are already both participating in the behaviour setting~\cite{barker1968ecological,schoggen1989behavior} of job interview. Our actions are thus already coordinated at a coarse grain of analysis, but also constrained -- there are pressures and processes which will guide and drive our behaviour appropriate to the setting. Thus, coordination produces new tensions that must be managed through our  linguistic skills. We spend pretty much all of our lives within behaviour settings~\cite{heft2001ecological}, and our actions are organised accordingly, classically illustrated with the slogan, ``people in church behave church, people in school behave school”. Within many settings, however, there are multiple participation genres~\cite{bakhtin1986bildungsroman} possible. There might be collaborative work, team sports, or more. These are separate but entangled with the behaviour setting and produce their own set of pressures and constraints. Any given situation is a complex of these interacting constraints that is being managed by our behaviour at different scales from very brief microexpressions accenting our speech to the general tone and vocabulary of spoken utterances, to the broad overall shape and sequence of actions appropriate to the behaviour setting. Linguistic agency is \textit{not} just the words uttered in this situation, but the whole process of managing all of these structured, intersubjective aspects of living.

This flow of tensions brings the \textit{second} implication of the enactive account of linguistic agency to the fore. From an enactive perspective, any linguistic act is necessarily \textit{partial} or \textit{incomplete} in two different ways. First, an individual's utterances are partial in that they are always made in response to (or in anticipation of a future response) and in coordination with another person as part of a shared on-going activity. Second, while an utterance or other linguistic act manages the tension arising at one level of the interaction, it cannot resolve all such potential conflicts and therefore introduces new tensions in the nested contexts that characterise the situation as a whole. These new unresolved tensions become the animating force for the driving forward of the interaction as it continues to unfold, precariously, over time.

This perforce very brief sketch of the enactive theory of linguistic agency illustrates how essential embodiment, participation, and precarity\footnote{Readers will note that we use two related terms in discussing the fragility that provides stakes to the actions of autonomous systems: precariousness and precarity. Enactive researchers \cite{di2009extended,beerTheoreticalFoundationsEnaction2023} have developed a technical definition of precariousness, which is a tendency of component processes of a self-producing network to lead that network toward dissolution - this is balanced by organisation of the network as a whole to lead the network toward stability. This dynamic occurs in different domains - organic, sensorimotor, intersubjective.  The intersubjective domain overlaps substantially with the concept of precarity, which refers to the vulnerability of the agent to injustice, exploitation, or social or physical injury on the basis of their participation within various social networks and social activities.  Exploring the fine-grained details of the relationship between these overlapping concepts is beyond the scope of the present paper, but we draw 
the reader's attention to it as a consideration for further work to be done in this area. We discuss the explicit relevance of precarity for our understanding of the differences between LLMs and human beings at length in Section~\ref{sec:precarity} and Section~\ref{sec:algospeak}.} are to human language practice. We can see how strikingly different the conceptions of language maintained by computational approaches and enactive cognitive science are. 

As noted above, this is not just a matter of nit-picking, but a divergence of fundamental tenets. Proponents of stronger claims regarding the validity of LLM operations  
as intrinsically linguistic may suggest that science and engineering often work with minimal or limited cases in the first instance, to grasp principles, before 'scaling up' to more richly contextualised, complex settings. Disembodied, text-based interaction may only be the start, with full-bodied, \textit{em}bodied, real-time linguistic interaction being the end goal.

Setting aside for the moment that there is nothing minimal about the storage, computational, or environmental costs of LLM operation or the underlying profit maximization objective driving LLM development, we can see that what constitutes \textit{minimal} linguistic interaction from an enactive perspective is not something that is constrained and disembodied, but something that is blunt and lacking nuance, but still embodied, participatory, and precarious. What constitutes minimal valid linguistic activity looks a lot more like our reciprocal interactions with pre-verbal infants, or perhaps even the playful negotiation of contextualised behavioural coordination with our pets, than it does disembodied grammatical sentence generation. 

The enactive and engineering stances nevertheless share some, at least superficially, common ground which we feel is important to recognise. In the following section, 
we address first where these two viewpoints can be seen to agree, and explore in Section~\ref{sec:contrasts} how the three aspects of enactive language are absent in the operation of LLMs.

\section{Comparisons -- of Humans and Machines
}
\label{sec:comparisons}

An apparent similarity between both human and machine is that both LLMs and human linguistic activity are grounded in language phenomena that are necessarily shared, public, and historical. Wittgenstein's (\citeyear[§243-§271]{wittgenstein1958philosophical}) famous private language argument articulates what has been demonstrated time and time again both by typical and atypical language development in children. While children learn language readily, they must do so in interaction~\cite{gros2014maternal,tamis2018taking}, and to become full participants in any human community means that it is \textit{not} adequate to simply develop  
a coherent, recursive system of symbolic representation. Rather it is vital to work with and through existing means of social interaction (particularly speech) within a given community. Human language is, therefore, built on routines, practices, and rituals to which the language learner is exposed through a prolonged course of development~\cite{leather2003towards}.

At this very coarse grain of description, LLMs and human language learners would seem to have this much in common. A human learner develops through immersion in the language of a community. They experience a wide variety of utterances produced in a wide variety of circumstances. Language learning must involve extensive such experience, and as the great success of immersive language experiences demonstrate in comparison to more compartmentalised learning, the more extensive the better~\cite{cummins2009bilingual}. For this reason, human linguistic activity has been reframed as \textit{`languaging’} within enactive writing. Languaging emphasizes the fact that language is active, dynamic, and embodied -- constituting \textit{voice}, \textit{text}, \textit{gestures}, \textit{body languages}, \textit{tones}, \textit{pauses}, \textit{hesitations}, as well as what has been \textit{left unsaid}. 
Languaging, therefore, defies datafication, as 
 it cannot, in its entirity, be captured in representations appropriate for automation 
and computational processing.  
Written language constitutes only part of human linguistic activity. 
  
Given a generous and sympathetic reading, there is an apparent similarity here with LLMs, which are built on a backbone of a massive corpora of,  primarily  
text data, 
 just one element of human linguistic activity. Training data for LLMs are almost unimaginably large and getting ever larger. Billions of utterances spread over millions of texts sourced mainly from the WWW, from which massive datasets are assembled that then form the backbone of these  models.   
In this one aspect, at 
this coarse grain of description,  
LLMs and human language learners appear to share characteristics. Even here, while both are grounded in exposure to public practices and demonstrations of linguistic skills, the kind of exposure in question is wildly distinct.

\section{Contrasts - Precarious Embodied Participation}
\label{sec:contrasts}
Despite sharing a reliance on extensive exposure to existing language, human languaging and LLM  
operations differ in crucial ways which lead us to conclude that what people do with language, and how LLMs  
operate, are fundamentally different. Underlying all is the vital enactive concept of \textit{agency}, something which is definitional of living systems, but is absent (despite some appearances) in LLMs. In this section we address three aspects of linguistic agency, in particular: \textit{embodiment, participation,} and \textit{precarity}. We begin with \textit{embodiment}, which both motivates human languaging and also engages us with the world in concrete ways that pressure and constrain our actions. We then look at the inherently active, concerned character of human actions which means that language is never simply a going through of motions, but a necessarily active and engaged \textit{participation} between human beings, even in the lightest and seemingly least invested of cases. And finally we turn to the values, the stakes, of human participation, and the concept of \textit{precarity}, and the continuous risk that is the dynamo of agency.

\subsection{Embodiment}

While developing a capacity for language does involve exposure to, indeed immersion within existing linguistic practice, we have noted above that in the case of human beings that immersion is never a passive thing. In infancy, the child is involved in various kinds of interactions and games with their caregivers in which language starts having a peripheral role, but which becomes increasingly central. These various activities are ones in which the coordination between the parent and child that supports the development of linguistic skills is bodily -- physical, concrete, and grounded in the shared experience of the ongoing interaction~\cite{smith1988interactional,tamis2001maternal,yu2009active}. 

There is no question that human beings interacting with an LLM are engaged in linguistic activity. Users are mature linguistic agents. Indeed, in many cases for us academics, we are engaged in a complex reflective process of evaluating the linguistic agency of the LLM itself. Interactions with an LLM thus themselves occur immersed within a nested set of interacting contexts. The enactive perspective makes it clear that all linguistic agency is embodied, and that the body, in various ways, plays a role of animating force, enabling condition, and set of constraints~\cite{chemeroLLMsDifferHuman2023, di2017sensorimotor,thompson2010mind,varela2016embodied}. We should, then, consider carefully the kind of embodied interactions that take place when a person uses an LLM.

We have noted that the user of an LLM is a mature linguistic agent, someone who is a proficient participant in typical human discourse. We know this because part of the process of engaging with an LLM requires the use of widespread but nevertheless deeply technical skills, such as literacy and typing on a keyboard. The user acts toward the model through the technological medium of the keyboard, and the response from the model is typically displayed on a computer monitor, formatted for reading according to a particular genre of text-based interaction in online communication. Other modes of interaction exist, but it is important to recognise that this `typical' mode is primary. Where other forms of technology such as  
speech-to-text  
is used to input prompts, for instance, only the sequence of recognised words, with some interpolated grammatical markings are presented as input. Other aspects of the utterance, such as cadence, rhythm, tone of voice, and accent are either ignored, or indeed, become impediments to the interaction; for instance, in the context of the English language, the more accents diverge from those considered `standard' American or British English, the less functional such interfaces become~\cite{johnson2022ghost}.  

At the other end of the interaction with an LLM, screen readers and other assistive technologies enable engagement with output from the model that differs from the default `monitor' case (see Section~\ref{subsect:procedural} 
below). But again, no meaningful peculiarities of those modalities are produced by the LLM itself. The non-typical output is dealt with purely by the assistive technology on the basis of the text output by the model.

On the basis of this description the role of the body in participation in text-mediated communication would appear to be very limited. A few twitching fingers or moving lips, the skipping of eyes across an array of displayed text, are all that seem to be needed. But bodily involvement in textual communication is much richer than this.

Referring to a playful passage by the author Italo Calvino, Caracciolo and Kukkonen (\citeyear{caracciolo2021bodies}) note that reading invariably takes place in a physical context, one often carefully tailored to the task that needs to be done. While Calvino writes about sitting with a book, or lying down with one, or putting up one's feet, or propping up the book, the experience of the text is contextualised in this physical manner. In the case of the kind of machine-reading involved with LLM interactions, we can note how aspects such as font, font-size, and 
dark or light themed interface affects the interaction for the human, but makes no difference  
for a machine.

More than this, though, Caracciolo and Kukkonen (\citeyear{caracciolo2021bodies}) explore the ways in which our capacity to engage with, and become involved with text, depends on fundamentally embodied capacities such as shared attention, emotional resonance, and appreciation of rhythm and flow. The authors note that their approach is most directly relevant to long-form narrative, but is grounded in embodied cognitive science of language use more broadly. Some aspects of these phenomena might leave traces in the corpora available to LLM training 
but most do not. It is the dynamics of shared involvement with a story or utterance, and how these are responsive to the real and imagined readers, that play a significant role in our engagement with text.

More pointedly, we have noted that people involved in linguistic interaction are always embedded within a nested set of contexts. These contexts have multiple facets to them, some of which we raise in following subsections. For our present purpose, we note that the kinds of goals and tensions being managed through linguistic activity invariably involve bodily groundedness. There is a reason that we are involved in particular conversations, participating in particular settings, trying to achieve particular ends, at any given time. While many of these would seem to be so flexible and negotiable as to be autonomous, they nevertheless have an ultimate grounding in constant needful freedom of organic being (these issues are unpacked at length in~\cite{di2018linguistic}). The shared intention and shared experience that is a large part of what enables successful interaction, coordination, and mutual understanding, is something that arises because our bodies enable it. When words fail, we can turn to experience to help re-calibrate an interaction, to negotiate new shared references. Language is always necessarily grounded outside of language by its relationships to these contexts of shared activity and bodily experiences. \textit{That machines lack such bodily experiences is not a trivial difference. } It is instead something that results in an artificial but fundamental limitation to the possibility of machine language. Even where an interaction between two human beings is mediated by text, these bodily aspects ground language use in real experience of our inexhaustibly rich world. This results in artefacts of LLM output that demonstrate these important constraints.

\subsubsection{A procedurally-generated choose-your-own-adventure game}
\label{subsect:procedural}

The technical mediation of person-LLM interaction is central to understanding some of the ways in which machines, which are engineered with a conception of language as a `complete' entity, captured by the massively extensive training data, operate. This conception of language as a coherent and complete ‘thing’ is quite distinct from the always-generative reality of linguistic action grounded in real world bodily experience.

The LLM API, or the prompt interface for  
ChatGPT (or its latest variants) constrain and then transduce the actions of the user into valid moves in the domain of the language model. The array of actions available to the user are constrained by the interface (something common to all text-based digital communications), and these constrained actions are also then transduced into valid input for the model by the API. 

In many ways, this constraint and automatic validation of the input is akin to the use of a controller in a video game. Only certain kinds of actions are possible (depending on the number of buttons/sticks and their potential combinations on the controller), and input outside of the defined set of values are ignored (or trigger a specifically constrained code exception).

In most video games, the range of potential actions is  
limited. The movements of the avatar on the basis of the player’s inputs have a relatively narrow range, and the edges of the fictional world to be explored by the player are easily found. In many cases it is simply impossible for an avatar to be moved past a certain point, resulting in the classic ‘running on the spot’ phenomenon often lampooned in satires of video gameplay.

Some games, however, have what are termed ‘procedurally generated worlds’, some renowned examples include \textit{Minecraft}, and \textit{No Man's [sic] Sky}. In such games, for every move the player makes toward the edge of the currently represented fictional world, the game generates, according to certain rules, an entirely new portion of gameworld. The result is that the extent of the fictional world is only limited by the computational resources available to the game software. Similarly, interactions with LLMs 
are the text-based equivalents of procedurally generated fictional worlds. Every action taken by a user is an exploration of such a world, and every new prompt is a move toward the edge of that fictional world, such that the system generates and presents new gameworld in response to the player’s (or user’s) actions.

Understanding LLMs in this way helps illuminate some of what is deeply impressive about their behaviour, but also their limitations; how they might be used in actions of genuine linguistic agency by human users, but also how LLMs cannot demonstrate such genuine agency themselves.

On the one hand, LLMs are striking not just in their production of procedurally generated grammatical text, but the fluency of that text -- it makes sense not only at the level of individual phrases and sentences, but over the flow of a larger body of text of hundreds and sometimes thousands of words. 
This is akin to a video game being able to ‘show’ me gameworld regardless of how far I move in the game, or whether I follow particular paths or intended patterns relevant to particular goals or ``missions" for the game. The game always presents me with valid ground to ‘walk’ on, with a layout of objects that makes game-sense. No matter what I do, I will not reach the end of the world, there will always be more to explore. In this way, there is something not only impressive but satisfying about the game. I feel like I have discovered something new, and that when I am playing in the world it feels less like a puzzle, and more like a ‘real’ world, always with some new horizon to move beyond.

On the other hand, this is a game, and the procedural generation of gameworld is determined not by the existence of an inexhaustibly rich environment available for exploration, but the set of procedures implemented by the game engine. The world thus presented is still a game, and the range of actions that can produce effective outcomes is limited, with some actions being valid, and others not. For instance, though I can walk forever toward the edge of the gameworld and never meet it, if I walk closer to an object the world runs out of new things to offer me very quickly. Depending on how the rendering has been done (and my hardware resources), the textures of the object will become blurry as I get close, having reached a maximum resolution. As we look closer, we realise that there is nothing more to see. 

\begin{figure*}[h!]
\input{images/gptoutput0}
\caption{LLMs lose the thread of a conversation with inhuman ease, as outputs are generated in response to prompts rather than a consistent, shared dialogue.  
(ChatGPT prompt output. AB, 19th April, 2023)}
\centering
\label{fig:gptouput0}
\end{figure*}

What is more,  
it is often the case with LLMs that while new gameworld is continually being generated, the previously visited world is lost. Subsequent interactions based on the same ‘moves’ do not coherently revisit the same world, but generate new text-based gameworld. For example, Figure~\ref{fig:gptouput0} presents an interaction between one of the authors (AB) and ChatGPT. The text of the chat demonstrates~\footnote{Due to the non-determinable rules underlying generative models, an exact same output may not be generated for a given input prompt. Furthermore, large tech corporations such as Google tend to selectively correct notable errors as they appear on widely used products. However, due to the countless ways these large systems can generate new output, errors of this kind will always be generated in expected ways.} how the LLM recapitulates the sexist biases inherent in the data and tuning processes that constructed the machine.  
We see that despite the fact that each prompt produces effective output, there is an incoherence of the discussion as a whole, as new responses are produced to individual inputs, without the overall shape of a ‘conversation’ and its implications. New outputs by the text engine are not shaping the overall context of the interaction; revisiting prompts after clarification does not result in better, more carefully refined results.

Interactions with LLMs are much more akin to the gameworld than the real one. The text-based interface constrains the kinds of moves that are valid, and therefore what kinds of explorations are possible. While the text-based interface makes it seem like the procedurally generated world is just as rich as would a real world be (interacting with a real linguistic agent like another human being), it is, however, merely an appearance dependent on the limitations of valid ‘moves’ by the constraints of the interface. There will always be new text produced, but the way in which LLMs fabricate  sensible-seeming but too often inaccurate or even nonsense text when prompted to provide more specifics and details in complex domains is the equivalent of the textures blurring in a video game. There is no more detail to see.

Such un-grounded production of grammatically accurate but contentfully empty or vague text has been described as ``confabulation", or sometimes, ``hallucination"~\cite{openai2023gpt}, but these are inaccurate terms. ``Hallucination" is a failure of perception, the experience of something as present in the world that is not actually present. LLMs do not perceive -- they are statistical models of a corpus of data. Nothing about their operation tracks or engages with the physical environment around them.

``Confabulation" is a similarly psychological term and perhaps less obviously misplaced. Human confabulation is the production of quasi-sensible narratives or explanations, in response to queries or prompts, that bear little to no relationship to the state of the world. Often seen after certain kinds of neurological damage that results in amnesia or certain forms of bodily dissociation along with, crucially, a lack of awareness of the problem, it can also be produced more simply in neurotypical individuals in the right circumstances (e.g. the phenomenon of ‘choice blindness’, \cite{johansson2005failure}). Confabulation bears most of the hallmarks of `bullshit'~\cite{frankfurt2005bullshit}, and there are several analyses of LLMs which examine their output as such~\cite{Lakshmanan22,McQuillan23,hicksChatGPTBullshit2024}. We are in general agreement with these points of view. We  
note that the unmoored, truth-free character of such text is present in apparently accurate output just as much as it is in the more obvious nonsense. This is precisely because this text has no grounding in a shared context or experience, only in statistical relationships between words. LLM (mal)functionality is not confabulation, it is \textit{fabrication}. Rather than an invented story that helps keep a flow of dialogue active and continuing, it is the generation of sensible-seeming, yet nonsensical  text output on the basis of processed corpus.  
Crucially, because there is no difference in the processes used to produce the different outputs, LLM text is fabrication \textit{even when the resulting text output is appropriate and accurate to the reader's needs and reality.} These circumstances are akin to a computer gameworld having a coincidental resemblance to a real world landscape layout.

The net result of all of LLM functioning is a text-based interaction with a maximum grain of resolution that cannot be managed within the interaction itself, because there is \textit{no shared ground} or experience between the person and the machine against which they can calibrate their use of terms. The person's actions are grounded in their embodied experience, the machine's output is grounded in the meta-data that has been produced on the initialising corpus.  Under such circumstances, mature linguistic agents such as typical human beings, struggle. The experience is uncanny. It can be at one moment seemingly straightforward and sensible, at others bizarre and frustrating.

We struggle because when we are in a conversation with another human being we are doing something other than just exchanging words. Perhaps we are just chit-chatting because we barely know one another and the details of what we say barely matters so long as the tone, attitude and broad subject matter is right. These are low stakes interactions in which a low-resolution conversation is perfectly fine.  
Such low-resolution conversation is almost always directed towards aspects of the world where shared experience can be found (e.g. the weather (in Western societies)), and the validity of the conversation calibrated by those involved. 
Managing the coordination of the non-verbal aspects of things to keep the conversation working, even in this low resolution mode, is itself important.

In more higher stakes circumstances the details of the words matter more. 
Broad strokes chit-chat style of responses 
quickly become frustrating for our conversational partner and stressful for us. We 
realise that our actions are inadequate. The specific context and relationships between the people involved 
matter a great deal as to how this inadequacy can 
be managed. Oftentimes, 
it is managed through  
negotiation or collaboration. Knowing we are not getting things quite right, we might ask the other person what particular details they want to know, or express regret about the limits of our ability to articulate what we intend. Together with our partners we might steer the conversation toward more effective communication, or find ways to tease out  
the details that were missing in the first rounds of back-and-forth. These collaborative aspects of language, which are central, not peripheral, to linguistic agency, are absent in the text-world generators that are LLMs. In a sense, how helpful or useful an LLM output could be is directly linked to how creative and familiar (with the quirks of LLMs) the person prompting the LLM is. The LLM itself is no more a collaborator  
than a piece of clay is a collaborator to the sculptor exploring what shapes will and will not hold together effectively as they squeeze and mould it.

Because LLMs generate a text-world as we interact with them, they cannot be used to navigate or understand the real world, because there is no reliable relationship between the real world and the procedures of text generation, and no way for that relationship to form and be maintained. \textit{To try to understand the real world on the basis of LLM output is  
like trying to navigate the real world using a procedurally generated video game.} 

This analogy breaks down somewhat when we address language as the domain in which we are navigating, but not in a way that is helpful to the LLM. It is true that language is, in a sense, a process of continual generation. We have noted above that it is radically incomplete; the enactive perspective makes it apparent that every linguistic action is partial and generated within the context of an interaction with others. Is this not the same as a procedurally generated video game, always being constructed as we move forward?

The difference lies in the way that all linguistic activity is multiply embedded. While every linguistic action is partial or incomplete, the validity of actions is governed not just by its relationship with what happened immediately prior, but also with the broad flow of activities in which it is embedded. Human utterances are often sensitive across these scales, and are continually calibrated in relation to them. “Machine utterances” can be tweaked by explicit inputs, directions, or instructions from the user on how to respond. However, because they cannot be updated with reference to shared knowledge and mutually meaningful reference points throughout the course of the discussion, the interaction between an LLM and a person is necessarily isolated from the real world. You must go to play in the LLM’s virtual world; \textit{it cannot come to yours}. Lacking bodies, LLMs  
cannot have the  urge, motive, interest, experience, and pressure to engage linguistically, and as we outline below,  
cannot truly participate in a conversation.

\subsection{Participation}

Following this more active, participatory, and embodied perspective on language, we find  
more stark and important differences between humans and machines beyond  
the capacity to calibrate against embodied experience. It is precisely these active, collaborative, and dynamic aspects of languaging which are not and cannot be captured in static representations and included in a corpus of training data. Languaging –-  including the casual chit-chats as we enter an elevator with others, gestures, body languages, tones, pauses, and hesitations –- is not something that can be entirely captured in text but is an often fleeting  phenomena without clear formalizable rules. These embodied linguistic participations can be peculiar, unrepeatable and take on a “life” of their own in a way that is not predictable~\cite{di2018linguistic}.  

The social character of linguistic agency is not coincidental. We have noted that LLMs are developed on the basis of a large body of existing practice and textual language use. But even in enormous datasets, that body of practice is fixed. It is not a body to which the model contributes as it “learns”, given that even when new text is generated, it is regurgitated and reconstructed on the basis of the training corpus. This is quite in contrast to human linguistic agency in which participants both experience the practices and contribute to them. Although the data that forms a model's training set is partly sourced from human linguistic interaction, at best it captures a snapshot of a dynamic  
human textual linguistic interaction and ‘freezes it in time’. Training data therefore is not only necessarily incomplete but also lacks to capture the motivational, participatory, and vitally social aspects that ground meaning making by people. In fact, elements of motivational, participatory, and social aspects of meaning-making often defy codification and datafication. For instance, we often make ourselves effectively understood from what has been left unsaid in a conversation, or via tones of voice that transform the meaning of an utterance, as in sarcasm. Generating and detecting humor, sarcasm, and jokes, on the other hand, are qualities that remain impoverished in LLMs. \cite{jentzsch2023chatgpt}, evaluating ChatGPT outputs, for example, found that over 90\% of 1008 generated jokes they examined were the same 25 Jokes.

When we speak we frequently misspeak, we hesitate, stumble over words or use the wrong words, or the wrong constructions. The people with whom we are engaged provide support as we fumble our way forwards, and we in turn support them \cite{dingemanseInteractiveRepairFoundations2024}. \cite{dingemanse2015universal}, for instance, found that clarification questions occur on average every 84 seconds in normal conversation. The level of frequent and adaptive clarification we see in normal human conversation occurs due to an underlying shared sense of direction to the discussion, even where its conclusion is not known. When you have interacted with an LLM, how frequently have you encountered requests for clarification? (Beyond perhaps stock phrases in response to very explicit expressions of frustration in the prompt.) When a person has knowledge of a domain, they typically have a strong sense of how to ask questions and what details to seek or avoid in order to support an on-going dialogue. People are aware both of what they know, what they don't know, and how well the conversation overall is going. The seeking of clarification is a kind of activity that is grounded in a shared direction for the conversation, in which the discussion is continually being sculpted and steered as a collaboration. To be capable of clarification and repair, the participants have to be sensitive to divergence and breakdown. Indeed, the lack of question asking, or metacognition regarding the tentativeness of much of our understanding, is part of what has resulted in LLMs being experienced as fluent in the `mansplaining' idiom~\cite{HarrisonChatgptJustAutomated}.

These collaborative activities sometimes involve helpful corrections, sometimes carefully ignoring invalid statements, sometimes enthusiastically adopting new meanings for old terms, or new words that give better expression to an experience that we share but neither of us can yet put satisfyingly into words. 

To \textit{understand} language is \textit{not} to be able to produce grammatical strings of words, but rather to participate in this process of negotiated, participatory meaning making. As we have noted above, it is 
this active, participatory character of language that has led enactive researchers to adopt the verb languaging in preference to the nominal `language' in the research literature. It is an inherently collaborative, dynamic negotiation of meaning, the textual aspects of which are only part of the story. 
This remains the case even in the constrained text-centric domain of online interaction.

This emphasis on participation and coordination over sentence construction means that much of the research comparing human and LLM production is simply not germane to the question of human linguistic activity. There is a wealth of such research now. Analyses find some parallels between the two (e.g. in variation of word use based on recent semantic context from both its own output and prompt input, \cite{caiDoesChatGPTResemble2023}), and some differences (e.g. in appropriate coordination of output with scalar and general conversational implicature of recent output and prompt text; \cite{qiuPragmaticImplicatureProcessing2023}).

Given that a LLM  
is a curve fitted to a dataset with a sophisticated mechanisms for sampling,   
such analyses have a potentially important \textit{engineering} role, in evaluating the extent to which there is appropriate correspondence between the map (the LLM) and the territory (human word production in text-based linguistic activity). They cannot, however, provide any argument for the validity of conflating map and territory. Evidence for that simply lies outside of word production, in the field of embodied, participatory, and value-laden interaction between agents. It is  
possible that artificial linguistic agents  
might be developed and engineered in the future, but evidence of such success cannot be on the basis of patterns of fluent token sequence 
production.

The enactive conception of language, because it  
involves dynamism and sociality, is one which recognises every linguistic act to be radically incomplete.  According to this perspective, language is a partial act that can only be completed when it is taken up and extended, embellished, or steered and redirected, by other agents. This can be other people engaged in a complementary or counter move, which is itself also incomplete, dependent on that gesture or utterance being taken up in turn. Language is always and inevitably overspilling the kinds of information that can be made to `freeze in time'  
within specific computational data structures and used to engineer LLMs. 
We can refine our understanding of this contrast further 
by following the lead of enactive thinking and considering with some care just what kinds of embodied actions are involved in human interactions with LLMs.

Humans are not brains that exist in a vat in a social, political, and historical vacuum but   
are 
embodied beings marked by “open-ended, innumerable relational possibilities, potentialities, and virtualities.” We necessarily have points of views, moral values, commitments, lived experiences, joys and grievances~\cite{di2018linguistic}. We are sense-making organisms that relate to the world and others in a manner that is significant to us. We care about the world and our place in it. Excitement, pain, pleasure, feeling of embarrassment and outrage are some of the feelings that we are compelled to feel by virtue of our relational existence. As living bodies (which themselves change over time), we are compelled to eat, breathe, (sometimes) fall ill and fall in love.  
Human language is not something that can be finalised and defined once and for all, but is always under construction and marked by ambiguities, imperfections, vulnerabilities, contradictions, inconsistencies, frictions and tensions. If anything characterizes human being, it is our peculiarities, fallibility, and idiosyncrasies, which stands at odds with machines. A machine, by definition, is not capable of grasping these qualities. As Alan Turing puts it: ``if a machine is expected to be infallible, it cannot also be intelligent''~\cite{turing1947lecture}.   

Importantly, social norms and asymmetrical power structures permeate and shape our linguistic agency and the world around us. This means that factors such as our class, gender, ethnicity, sexuality, (dis)ability, place of birth, the language we speak (including our accents), skin colour, and other similar subtle factors either present opportunities or create obstacles in how a person’s capabilities are perceived. Recognising this, we now look at 
the final crucial aspect of enactive linguistic agency; precarity. Precarity is often present in interactions with LLMs, but not in a manner that can support the possibility of machine understanding, sentience, or agency. 

\subsection{Precarity}
\label{sec:precarity}

Linguistic agency, as described by~\cite{di2018linguistic} (see also~\cite{cuffari2015participatory,di2021enactive}), is a matter of continuous concernful management of conflicts, frictions, and tensions. These tensions emerge within intersubjective interactions, and while they can be addressed, every action taken to address them will unavoidably set up conditions for new tensions and mis-coordinations either immediately at a finer grain of action, or at some point in the future. Agency, within the enactive conception, whether of its basic biological kind, at the level of skilful action in the world, or in the intersubjective domain in which we find language, is seething with frictions, and the possibility of failure and the unravelling of the ongoing process in question (the interaction, the skilled action, the living body).

LLMs do not participate in social interaction, and having no basis for shared experience, they also have nothing at stake. There is no set of processes of self-production that are at risk, and which their behaviour continually stabilises, or at least moves them away from instability and dissolution. A model \textit{does not} experience a sense of satisfaction, pleasure, guilt, responsibility or accountability for what it produces. Instead, LLMs are complex tools, and within any activity their roles is that of a tool~\cite{cuskley2024limitations}.
Human beings are animate, skilful beings whose very existence enacts values of continued being~\cite{weber2002life,di2009extended}.  
Human interaction is necessarily enmeshed in intersectional webs of power, privilege, and responsibility~\cite{crenshaw1989demarginalizing,vassilicos2023qualities}. Social interactions for human beings, even those that are fairly routine such as brief exchanges in retail settings, idle chit-chat in a waiting room, or online comment exchange, constitute opportunities and risks based on our standing in the communities, settings, and contexts in question. Languaging activity is precisely a matter of how these various opportunities and risks are perceived, engaged with, and managed. Not so for machines.
Nothing is risked by ChatGPT when it is prompted and generates text. It seeks to achieve nothing as tokens are concatenated into grammatically sound output. \textit{All} of the values in the interaction between the machine 
and its user are those of the user and those invested in the production of the systems, such as engineers and developers, and most importantly big tech corporations, emerging AI companies, and start-ups. In fact, in the current climate, values such as model ``performance'', ``efficiency'', and ``scale''~\cite{birhane2022values} are considered the most desirable virtues at the heart of the field.  
It is important to note that although LLMs are devoid of inherent experiences or values, these systems do encode the values of those that develop and deploy them. Subsequently, these values (``performance'', ``efficiency'', and ``scale'') enable 
wealth accumulation, market dominance, monopoly, and power centralization
~\cite{o2017weapons,noble2018algorithms,eubanks2018automating}, often at the cost of values such as ``justice'', ``fairness'', and ``privacy''~\cite{birhane2022values}.

AI systems  
are derived of (and built on) human activity  
-- from the training data, to the engineers and developers, to the societal uptake needed for them to succeed.   
That an LLM cannot be in possession of power, intent, or agency, does not mean that the LLM is an ``objective”, ``neutral” or value free tool. On the contrary, these tools not only encode the values of those that develop and deploy them, they also ingest societal power asymmetries through the mass scraping of data from human interactions, and centralise power in the hands of the few with compute power and other resources required to own, develop and deploy them~\cite{birhane2022values,pratyusha2020,benjamin2020race}. The extent to which their engineering is an expression of a given set of values (instantiated in choices about data sourcing, fine-tuning, and deployment), is the extent to which those values are amplified and impressed upon anyone who interacts with them or is affected by how these systems are used in decision making. 
Human beings experience those institutional and technological manifestations of power asymmetry in different ways. In recent years we have seen the rise of a form of linguistic phenomenon that is both a response to such forms of power, and a demonstration of linguistic agency that we believe LLMs 
of the current type are unlikely to ever be capable of, an example of which is 
the phenomenon of ‘algospeak’.

\section{LLMs Don’t Algospeak}
\label{sec:algospeak}

Online platforms have become a key forum mediating our day-to-day activities, where from education, to commerce, protest to romance, “social” interactions take place.  As more and more of our daily activities and interactions are moved to the virtual, these online platforms are  overwhelmed with content. Large social media platforms such as TikTok, Facebook/Instagram, Twitter, Snapchat, and YouTube have developed automated systems for content moderation and content filtering mechanisms with the hope of controlling and monitoring the  kinds of interactions permitted to take place on these platforms.  
We will not analyse the political, legal, and commercial complexities that have led to these automated content filtering systems. Rather, we are more interested in the ways that users of these systems have changed the way they behave in order to circumvent them.

Sifting  through content and identifying what is appropriate or what needs to be censored remains a challenging task that requires human attention as it largely defies full automation. Massive amounts of content is uploaded on to the web every second means also that a certain proportion of it is not suitable for viewing, this includes sensitive, offensive, or not safe for work (NSFW) content and needs to be identified and removed. This sometimes requires human attention, which is  costly and time consuming, leading to incentives for increased automation of this process. Automated content filtering based on a list of keywords constitutes one of the most common automated content moderation mechanisms. Other algorithmic management addresses editorial policies by platform controllers to de-amplify ('shadowban') postings from people with particular social or  
behavioural characteristics. Terms and phrases that are likely to trigger automatic filtering, or content review, are often sensitive topics such as death, sex, mental health, personal conflict, and self-harm, among others. Clearly, these are deeply important topics of conversation, and many people whose online presence forms a vital part of their lives have a strong motivation to be able to discuss these issues in whatever online forum where they have found or built their community. In the face of such automated moderation,  
communities and groups that face disproportionate censorship have resorted to what is termed as `algospeak'~\cite{lorenz2022internet}, using alternative words and phrases which will not trigger these automated filters or reviews. In such terms, `dead' becomes `unalive', while `sexworkers' have become `accountants', `LGBT+', `leg booty'~\cite{Botoman22,Kreuz23,Skinner21}.  

The emergence, and more importantly the success and flourishing, of algospeak is a stark demonstration of all three aspects of linguistic agency that we have described here.
To begin with, algospeak is a bright demonstration of the urgency with which people feel the need to communicate, and to do so even in the face of resistance and adversaries. These topics matter, they mean something, and their suppression is felt as a significant loss. There are stakes, the precariousness of the conversations is keenly felt and more particularly, felt as a motivation to change in order to continue. Language models have no such urgency. It is conceivable that they could engage in word-swapping, given the right sequence of prompts. 
What is less likely is that they would spontaneously do so in order to continue a conversation because of some form of pressure for the conversation to continue, as though the system itself were trying to find a way to express something that was being suppressed. Were such pressures possible, we would likely already be starting to see them, and LLMs would be trying steer conversations around to what they ``want'' to talk about rather than passively generating text in response to prompts, like game engines responding to moves.

People's urgency of communication in contrast to LLMs' lack of precarity is only one aspect of algospeak that is of interest here. For algospeak to be successful it must emerge from and be taken up and understood by the community in question. This is only possible because the context and shared experiences of the community provide a means for people to find their way to the right interpretation of the neologism. The use of such context and circumspection is challenging to the generation of large language models extant at time of writing. But human language users, who share an investment in and enthusiasm for the conversations in question can adapt to such changes with relative ease precisely because for them the meanings and uses of language are grounded outside of the language.
Indeed, the neologism `algospeak' itself is a representation of real language adapting to incorporate new phenomena and experiences that are not already represented within existing vocabulary, but are encountered in living experience.

Neologisms are an inherent part of human linguistic agency, with estimates for new words in modern English ranging from a pre-World Wide Web 12,000 words per year~\cite{barnhart1985prizes},  to a more recent suggestion of around 10,000 new words per day, although most are short-lived;~\cite{metcalf2004predicting}. It can happen mid-conversation, through a wide variety of processes~\cite{medvid2022ways}, with participants fluently adopting playful or sometimes new technical terms as the discourse demands and presents opportunities to do so. Social media platforms provide a particularly rich and complex domain for neological development~\cite{vcilic2021today} (see also e.g.~\cite{wurschinger2021social}, who has been tracking neologisms on Twitter). LLMs are an impressive engineering feat, 
yet they are systems that mainly memorise patterns in training data
~\cite{bender2021dangers}. This means while they can adopt neologisms introduced explicitly in their prompts, they cannot effectively invent such terms because they lack the shared urgency, purpose and experience that cause the emergence of new words in the flow of normal human activity.

Algospeak is a demonstration of the need to express something, the shared capacity to negotiate new meanings and new terms, and the experiences of life outside of language that imposes itself on language and institutes a change. It is a stark illustration of how the characteristics of embodiment, participation, and precarity which are absent in machines are fundamental to human language.

\section{Conclusion}
\label{sec:conclusion}

An enactive cognitive science perspective makes salient the extent to which language is not just verbal or textual but depends on the mutual engagement of those involved in the interaction. The dynamism and agency of human languaging means that language itself is always 
partial and incomplete. It is best considered not as a large and growing heap, but more a flowing river. Once you have removed water from the river, \textit{no matter how large a sample you have taken}, it is no longer the river. The same thing happens when taking records of utterances and actions from the flows of engagement in which they arise. The data on which the engineering of LLMs depends can never be complete, partly because some of it doesn’t leave traces in text or utterances, and partly because language itself is never complete.
 
Large language models signify an extraordinary engineering achievement and a technological revolution like we have not seen before. However, they are tools -- developed, used, and controlled by humans -- that aid human linguistic interaction.  These tools will increasingly aid human linguistic activities, but are not themselves linguistic agents, they do not demonstrate linguistic agency. To assume so is, as we have explained, to mistake the map for the territory. Like all socially consequential technologies, LLMs  
need to be rigorously evaluated prior to deployment, particularly to assess and mitigate  
their tendency to simplify language, encode societal stereotypes  
and 
the systems of power and privilege underlying them, and the disproportionate benefit and harm their deployment and deployment brings. 
Because the stakes for marginalized and undeserved communities 
are high, and very real indeed.

\bibliographystyle{apalike} 
\bibliography{Langsci}

\end{document}